%% file: main.tex
\documentclass[11pt]{article} % For LaTeX2e
\usepackage{url}
\usepackage{smile}
\usepackage{graphicx} % more modern
\usepackage{algorithm}
\usepackage{algorithmic}
\usepackage{epstopdf}
\usepackage{wrapfig}
\usepackage[colorlinks, linkcolor=blue, anchorcolor=blue, citecolor=blue]{hyperref}
\usepackage[margin=1in]{geometry}
\usepackage[normalem]{ulem}
\usepackage[export]{adjustbox}% http://ctan.org/pkg/adjustbox
\usepackage{mathtools, cuted}
\usepackage{natbib}
\usepackage{enumerate}
\usepackage{enumitem}

\usepackage{kpfonts}
\DeclareMathAlphabet{\mathsf}{OT1}{cmss}{m}{n}

\SetMathAlphabet{\mathsf}{bold}{OT1}{cmss}{bx}{n}

\newtheorem*{theorem*}{Theorem}

\title{\huge \bf Transformer Hawkes Process\footnote{Published as a conference paper in ICML 2020.}}

\author{Simiao Zuo, Haoming Jiang, Zichong Li, Tuo Zhao and Hongyuan Zha \footnote{Zuo, Jiang and Zhao are affiliated with Georgia Tech, Li is affiliated with University of Science and Technology of China, and Zha is affiliated with Shenzhen Research Institute of Big Data,
The Chinese University of Hong Kong (currently on leave from Georgia Tech). Correspondence to \url{simiaozuo@gatech.edu}, \url{tourzhao@gatech.edu}, \url{zhahy@cuhk.edu.cn}.}}

\newcommand{\commentout}[1]{}

\begin{document}

\maketitle

\input{abstract.tex}
\input{introduction.tex}
\input{background.tex}
\input{model.tex}
\input{experiment.tex}
\input{conclusion.tex}

\bibliography{main}
\bibliographystyle{ims}
\end{document}

%% file: abstract.tex
%!TEX root = main.tex

\begin{abstract}

Modern data acquisition routinely produce massive amounts of event sequence data in various domains, such as social media, healthcare, and financial markets. These data often exhibit complicated short-term and long-term temporal dependencies. However, most of the existing recurrent neural network based point process models fail to capture such dependencies, and yield unreliable prediction performance. To address this issue, we propose a Transformer Hawkes Process (THP) model, which leverages the self-attention mechanism to capture long-term dependencies and meanwhile enjoys computational efficiency. Numerical experiments on various datasets show that THP outperforms existing models in terms of both likelihood and event prediction accuracy by a notable margin. Moreover, THP is quite general and can incorporate additional structural knowledge. We provide a concrete example, where THP achieves improved prediction performance for learning multiple point processes when incorporating their relational information.

\end{abstract}

%% file: introduction.tex
%!TEX root = main.tex

\section{Introduction}
Event sequence data are naturally observed in our daily life. Through social media such as Twitter and Facebook, we share our experiences and respond to other users’ information \citep{hawkes-friendship}. In these websites, each user has a sequence of events such as tweets and interactions. Hundreds of millions of users generate large amounts of tweets, which are essentially sequences of events at different time stamps. Besides social media, event data also exist in domains like financial transactions \citep{hawkes-finance} and personalized healthcare \citep{hawkes-health}. For example, in electronic medical records, tests and diagnoses of each patient can be treated as a sequence of events.
Unlike other sequential data such as time series, event sequences tend to be asynchronous \citep{stochastic}, which means time intervals between events are just as important as the order of them to describe their dynamics. Also, depending on specific application requirements, event data show sophisticated dependencies on their history.

Point process is a powerful tool for modeling sequences of discrete events in continuous time, and the technique has been widely applied. Hawkes process \citep{hawkes-process, sp} and Poisson point process are traditionally used as examples of point processes. However, the simplified assumptions of the complicated dynamics of point processes limit the models' practicality. As an example, Hawkes process states that all past events should have positive influences on the occurrence of current events. However, a user on Twitter may initiate tweets on different topics, and these events should be considered as unrelated instead of mutually-excitepd.

To alleviate the over-simplifications, likelihood-free methods \citep{likelihood-free, likelihood-free-rl} and non-parametric models like kernel methods and splines \citep{spline} have been proposed, but the increasing complexity and quantity of collected data crave for more powerful models. With the development of neural networks, in particular deep neural networks, focuses have been placed on incorporating these flexible models into classical point processes. Because of the sequential nature of event steams, existing methods rely heavily on Recurrent Neural Networks (RNNs). Neural networks are known for their ability to capture complicated high-level features, in particular, RNNs have the representation power to model the dynamics of event sequence data. In previous works, either vanilla RNN \citep{rnn-hawkes} or its variants \citep{neural-hawkes, rnn-pp} have been used and significant progress in terms of likelihood and event prediction have been achieved.

However, there are two significant drawbacks with RNN-based models. First, recurrent neural networks, even those equipped with forget gates, such as Long Short-Term Memory \citep{LSTM} and Gated Recurrent Units \citep{GRU}, are unlikely to capture long-term dependencies. In financial transactions, short-term effects such as policy changes are important for modeling buy-sell behaviors of stocks. On the other hand, because of the delays in asset returns, stock transactions and prices often exhibit long-term dependencies on their history. As another example, in medical domains, at times we are interested in examining short-term dependencies on symptoms such as fever and cough for acute diseases like pneumonia. But for certain types of chronic diseases such as diabetes, long-term dependencies on disease diagnoses and medications are more critical. Desirable models should be able to capture these long-term dependencies. Yet with recurrent structures, interactions between two events located far in the temporal domain are always weak \citep{long-term-depend}, even though in reality they may be highly correlated. The reason is that the probability of keeping information in a state that is far away from the current state decreases exponentially with distance.

The second drawback is trainability of recurrent neural networks. Training deep RNNs (including LSTMs) is notoriously difficult because of gradient explosion and gradient vanishing \citep{rnn-difficult}. In practice, single-layer and two-layer RNNs are mostly used, and they may not successfully model sophisticated dependencies among data \citep{depen-difficult}. Additionally, inputs are fed into the recurrent models sequentially, which means future states must be processed after the current state, rendering it impossible to process all the events in parallel. This limits RNNs' ability to scale to large problems.

Recently, convolutional neural network variants that are tailored for analyzing sequential data \citep{wavenet, conv-s2s, conv-compare} have been proposed to better capture long-term effects. However, these models enforce many unnecessary dependencies. This particular downside plus the increased computational burdens deem these models insufficient.

\begin{figure}[t!]
    \centering
    \includegraphics[width=0.5\linewidth]{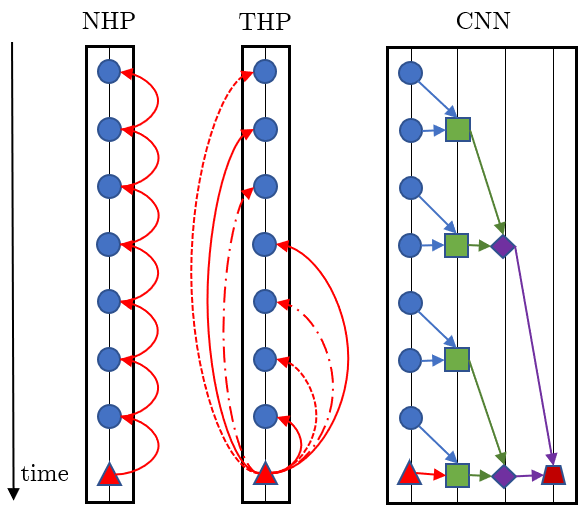}
    \caption{Illustration of dependency computation between the last event (the red triangle) and its history (the blue circles). RNN-based NHP models dependencies through recursion. THP directly and adaptively models the event's dependencies on its history. Convolution-based models enforce static dependency patterns.}
    \label{fig:encoding}
\end{figure}

To address the above concerns, we propose the Transformer Hawkes Process (THP) model that is able to capture both short-term and long-term dependencies whilst enjoying computational efficiency. Even though the Transformer \citep{transformer} is widely adopted in natural language processing, it has rarely been used in other applications. We remark that such an architecture is not readily applicable to event sequences that are defined in a continuous-time domain. To the best of our knowledge, our proposed THP is the first of this type in point process literature.

Building blocks of THP are the self-attention modules \citep{attention}. These modules directly model dependencies among events by assigning attention scores. A large score between two events implies a strong dependency, and a small score implies a weak one. In this way, the modules are able to adaptively select events that are at any temporal distance from the current event. Therefore, THP has the ability to capture both short-term and long-term dependencies. Figure \ref{fig:encoding} demonstrates dependency computation of different models.

The non-recurrent structure of THP facilitates efficient training of multi-layer models. Transformer-based architectures can be as deep as dozens of layers \citep{bert,gpt2}, where deeper layers capture higher order dependencies. The ability to capture such dependencies creates models that are more powerful than RNNs, which are often shallow. Also, THP allows full parallelism when calculating dependencies across all events, i.e., the computation between any two event pairs is independent with each other. This yields a model presenting strong efficiency.

Our proposed model is quite general, and can incorporate additional structural knowledge to learn more complicated event sequence data, such as multiple point processes over a graph. In social networks, each user has her own sequence of events, like tweets and comments. Sequences among users can be related, for example, a tweet from a user may trigger retweets from her followers. We can use graphs to model these follower-followee relationships \citep{social-learning, coevolve}, where each vertex corresponds to a specific user and each edge represents connections between the two associated users. We propose an extension to THP that integrates these relational graphs \citep{network-science, hawkes-net} into the self-attention module via a similarity metric among users. Such a metric can be learned by our proposed graph regularization.

We experiment THP on five datasets to evaluate both validation likelihood and event prediction accuracy. Our THP model exhibits superior performance to RNN-based models in all these experiments. We further test our structured-THP on two additional datasets, where the model achieves improved prediction performance for learning multiple point processes when incorporating their relational information. Our code is available at \url{https://github.com/SimiaoZuo/Transformer-Hawkes-Process}.

The rest of this paper is organized as follows:  Section~\ref{sec:background} introduces the background; Section~\ref{sec:THP} introduces our proposed transformer Hawkes process model; Section~\ref{sec:structure-thp} demonstrates an extension of our model to multiple event sequences on graphs; Section~\ref{sec:experiment} presents numerical experiments on various real datasets; Section~\ref{sec:conclusion} draws a brief conclusion.

%% file: background.tex
%!TEX root = main.tex

\section{Background} \label{sec:background}

We briefly review Hawkes Process \citep{hawkes-process}, Neural Hawkes Process \citep{neural-hawkes}, and Transformer \citep{transformer} in this section.

\vspace{0.05in}

\noindent $\bullet$ \textbf{Hawkes Process} is a doubly stochastic point process, whose intensity function is defined as
\begin{equation} \label{eq:inten-hp}
    \lambda(t) = \mu + \sum_{j: t_j<t} \psi (t-t_j).
\end{equation}
Here $\mu$ is the base intensity and $\psi(\cdot)$ is a pre-specified decaying function, i.e., exponential function and power-law function. Intuitively, Eq.~\ref{eq:inten-hp} means that each of the past events has a positive contribution to occurrence of the current event, and this influence decreases through time. However, a major limitation of this formulation is the simplification that history events can never inhibit occurrence of future events, which is unrealistic in complex real-life scenarios.

\vspace{0.05in}

\noindent $\bullet$ \textbf{Neural Hawkes Process} generalizes the classical Hawkes process by parameterizing its intensity function with recurrent neural networks. Specifically,
\begin{equation} \nonumber
    \begin{split}
        & \lambda(t) = \sum_{k=1}^K \lambda_k(t) = \sum_{k=1}^K f_k \big(\mathbf{w}_k^\top \mathbf{h}(t) \big), \quad t \in (0,T], \quad \text{where }
        f_k(x) = \beta_k \log \Big( 1+ \exp\Big(\frac{x}{\beta_k} \Big) \Big).
    \end{split}
\end{equation}
Prediction is then $\mathbb{P}[k_t=k] = \lambda_k(t) / \lambda(t)$. Here, $\lambda(t)$ is the intensity function, $K$ is the number of event types, and $\mathbf{h}(t)$s are the hidden states of the event sequence, obtained by a continuous-time LSTM (CLSTM) module. CLSTM is an interpolated version of the standard LSTM, and it allows us to generate outputs in a continuous-time domain. Also, $f_k(\cdot)$ is the softplus function with parameter $\beta_k$ that guarantees a positive intensity. One downside of the neural Hawkes process is that intrinsic weaknesses of RNNs are inherited, namely the model is unable to capture long-term dependencies and is difficult to train.

\vspace{0.05in}

\noindent $\bullet$ \textbf{Transformer} is an attention-based model that has been broadly applied in tasks such as machine translation \citep{bert} and language modeling \citep{gpt2}. Despite its success in natural language processing, it has rarely been used in other areas. We remark that the Transformer architecture is not directly applicable to model point processes. In particular, time intervals between any two events can be arbitrary in event streams, while in natural languages, words are observed on regularly spaced time intervals. Therefore, we need to generalize the architecture to a continuous-time domain.

% \textbf{Hawkes Process} \citep{hawkes-process} is a doubly stochastic point process, whose intensity function is defined as
% \begin{equation} \label{eq:inten-hp}
%     \lambda(t) = \mu + \sum_{j: t_j<t} \psi (t-t_j).
% \end{equation}
% Here $\mu$ is the base intensity and $\psi(\cdot)$ is a pre-specified decaying function, i.e., exponential function \citep{hp-exp} and power-law function \citep{hp-power}. Intuitively, Eq.~\ref{eq:inten-hp} means that each of the past events has a positive contribution to occurrence of the current event, and this influence decreases through time. However, a major limitation of this formulation is the simplification that history events can never inhibit occurrence of future events, which is unrealistic in complex real-life scenarios.

%% file: model.tex
%!TEX root = main.tex

\section{Model} \label{sec:THP}

We introduce our proposed Transformer Hawkes Process. Suppose we are given an event sequence $\mathcal{S}=\{(t_j, k_j)\}_{j=1}^L$ of $L$ events, where each event has type $k_j \in \{1,2,\hdots,K\}$, with a total number of $K$ types. Then each pair $(t_j,k_j)$ corresponds to an event of type $k_j$ occurs at time $t_j$.

%%%%%%%%%%%%%%%%%%%%%%%%%%%%%%%%%%%%%%%%
\subsection{Transformer Hawkes Process} \label{sec:transformer-arch}

\begin{figure}[b!]
    \centering
    \includegraphics[width=0.5\linewidth]{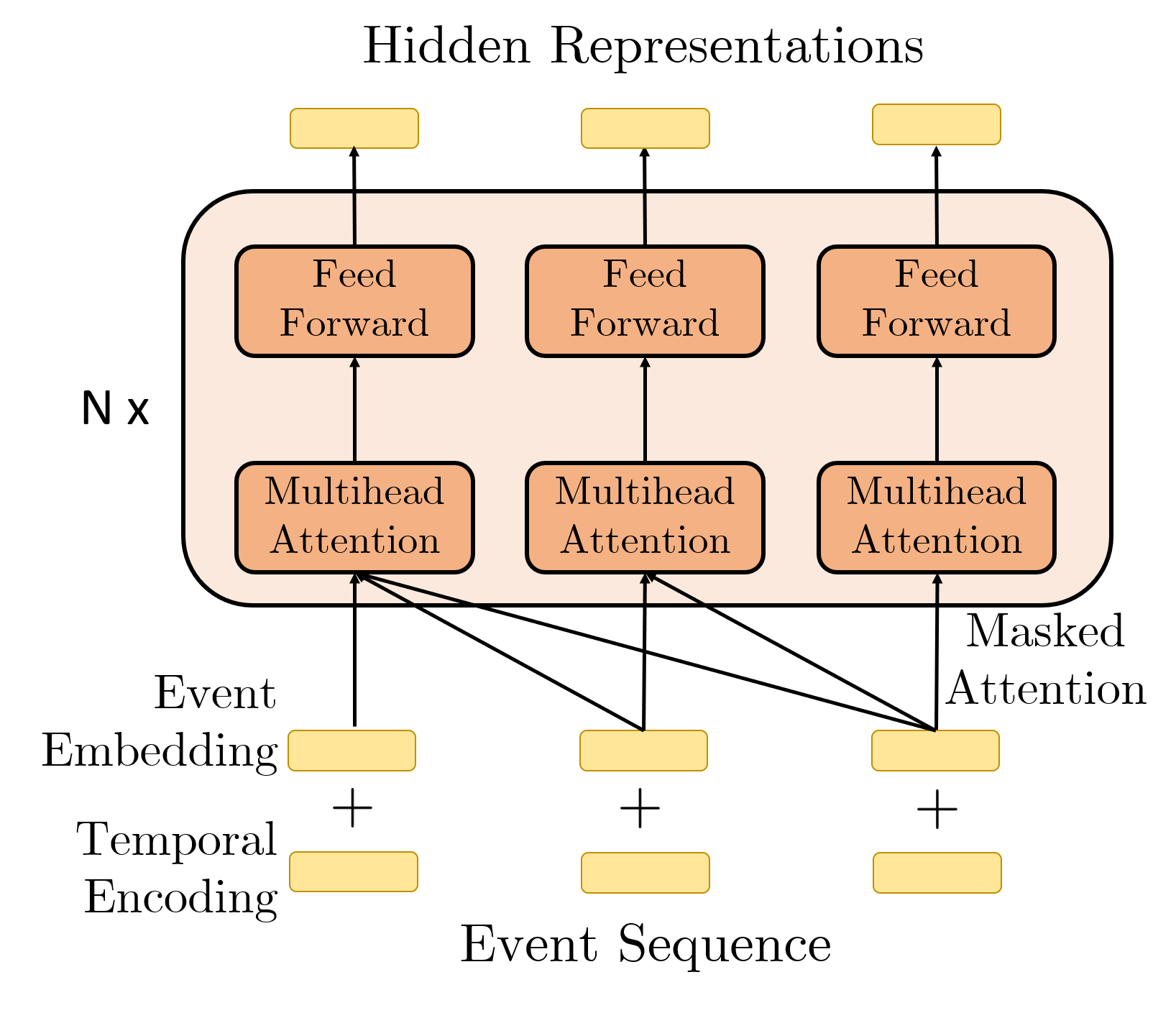}
    \vskip -0.2in
    \caption{Architecture of the Transformer Hawkes Process. Each event sequence $\mathcal{S}$ is fed through embedding layers and $N$ multi-head self-attention modules. Outputs of the THP are hidden representations of events in $\mathcal{S}$, with history information encoded.}
    \label{fig:THP-arch}
\end{figure}

The key ingredient of our proposed THP model is the self-attention module. Different from RNNs, the attention mechanism discards recurrent structures. However, our model still needs to be aware of the temporal information of inputs, i.e., time stamps. Therefore, analogous to the original positional encoding method \citep{transformer}, we propose to use a temporal encoding procedure, defined by
\begin{equation} \label{eq:temp-encoding}
    [\mathbf{z}(t_j)]_i = 
    \begin{cases}
        \cos \big( t_j / 10000^\frac{i-1}{M} \big), & \text{if $i$ is odd}, \\[2pt]
        \sin \big( t_j / 10000^\frac{i}{M} \big),  & \text{if $i$ is even}.
    \end{cases}
\end{equation}

Eq.~\ref{eq:temp-encoding} uses trigonometric functions to define a temporal encoding for each time stamp, i.e., for each $t_j$, we deterministically computes $\mathbf{z}(t_j) \in \mathbb{R}^M$, where $M$ is the dimension of encoding. Other temporal encoding methods can also be applied, such as the relative position representation model \citep{relative-encoding}, where two temporal encoding matrices are learned instead of pre-defined.

Besides temporal encoding, we train an embedding matrix $\mathbf{U} \in \mathbb{R}^{M \times K}$ for the event types, where the $k$-th column of $\mathbf{U}$ is a $M$-dimensional embedding for event type $k$. For any event of type $k_j$, let $\mathbf{k}_j$ be its one-hot encoding (a $K$-dimensional vector with all $0$s except for the $k_j$-th index, which has value $1$), then its embedding is $\mathbf{U}\mathbf{k}_j$. Notice that for any event and its corresponding time stamp $(t_j, k_j)$, the temporal encoding $\mathbf{z}(t_j)$ and the event embedding $\mathbf{U}\mathbf{k}_j$ both reside in $\mathbb{R}^M$. Embedding of the event sequence $\mathcal{S}=\{(t_j, k_j)\}_{j=1}^L$ is then specified by
\begin{equation} \label{eq:embedding}
    \mathbf{X} = \big( \mathbf{U}\mathbf{Y} + \mathbf{Z} \big)^\top,
\end{equation}
where $\mathbf{Y}=[\mathbf{k}_1, \mathbf{k}_2, \hdots, \mathbf{k}_L] \in \mathbb{R}^{K\times L}$ is the collection of event type one-hot encodings, and $\mathbf{Z}=[\mathbf{z}(t_1),$ $\mathbf{z}(t_2), \hdots, \mathbf{z}(t_L)] \in \mathbb{R}^{M\times L}$ is the concatenation of event time encodings. Notice that $\mathbf{X} \in \mathbb{R}^{L\times M}$ and each row of $\mathbf{X}$ corresponds to the embedding of a specific event in the sequence.

After the initial encoding and embedding layers, we pass $\mathbf{X}$ through the self-attention module. Specifically, we compute the attention output $\mathbf{S}$ by
\begin{equation} \label{eq:attention}
    \begin{split}
        \mathbf{S} &= \mathrm{Softmax} \left( \frac{\mathbf{Q}\mathbf{K}^\top}{\sqrt{M_K}} \right) \mathbf{V},
        \quad \text{where }
        \mathbf{Q} = \mathbf{X} \mathbf{W}^Q, \
        \mathbf{K} = \mathbf{X} \mathbf{W}^K, \
        \mathbf{V} = \mathbf{X} \mathbf{W}^V.
    \end{split}
\end{equation}
Here $\mathbf{Q}$, $\mathbf{K}$, and $\mathbf{V}$ are the query, key, and value matrices obtained by different transformations of $\mathbf{X}$, and $\mathbf{W}^Q, \mathbf{W}^K \in \mathbb{R}^{M\times M_K}, \mathbf{W}^V \in \mathbb{R}^{M\times M_V}$ are weights for the linear transformations, respectively. In practice using multi-head self-attention to increase model flexibility is more beneficial for data fitting. To facilitate this, different attention outputs $\mathbf{S}_1, \mathbf{S}_2, \hdots, \mathbf{S}_H$ are computed using different sets of weights $\{\mathbf{W}^Q_h,\mathbf{W}^K_h,\mathbf{W}^V_h\}_{h=1}^H$. The final attention output for the event sequence is then
\begin{equation} \nonumber
    \mathbf{S} = \big[ \mathbf{S}_1, \mathbf{S}_2, \hdots, \mathbf{S}_H \big] \mathbf{W}^O,
\end{equation}
where $\mathbf{W}^O \in \mathbb{R}^{HM_V \times M}$ is an aggregation matrix.

We highlight that the self-attention module is able to directly select events whose occurrence time is at any distance from the current time. The $j$-th column of the attention weights $\text{Softmax}(\mathbf{Q}\mathbf{K}^\top/\sqrt{M_K})$ signifies event $t_j$'s extent of dependency on its history. In contrast, RNN-based models encode history information sequentially via hidden representations of the events, i.e., the state of $t_j$ depends on that of $t_{j-1}$, which in turn depends on $t_{j-2}$, etc. Should any of these encodings be weak, i.e., the RNN fails to learn sufficient relevant information for event $t_k$, hidden representations of any event $t_j$ where $j \geq k$ will be inferior.

The attention output $\mathbf{S}$ is then fed through a position-wise feed-forward neural network, generating hidden representations $\mathbf{h}(t)$ of the input event sequence:
\begin{equation} \label{eq:hidden-rep}
    \begin{split}
        &\mathbf{H} = \mathrm{ReLU} \big( \mathbf{S} \mathbf{W}^{\text{FC}}_1 + \mathbf{b}_1 \big) \mathbf{W}^{\text{FC}}_2 + \mathbf{b}_2, \quad
        \mathbf{h}(t_j) = \mathbf{H}(j,:) . 
    \end{split}
\end{equation}
Here $\mathbf{W}^{\text{FC}}_1 \in \mathbb{R}^{M \times M_H}$, $\mathbf{W}^{\text{FC}}_2 \in \mathbb{R}^{M_H \times M}$, $\mathbf{b}_1 \in \mathbb{R}^{M_H}$, and $\mathbf{b}_2 \in \mathbb{R}^M$ are parameters of the neural network, and $\mathbf{W}^{\text{FC}}_2$ has identical columns. The resulting matrix $\mathbf{H} \in \mathbb{R}^{L \times M}$ contains hidden representations of all the events in the input sequence, where each row corresponds to a particular event.

To avoid ``peeking into the future'', our attention algorithm is equipped with masks. That is, when computing the attention output $\mathbf{S}(j,:)$ (the $j$-th row of $\mathbf{S}$), we mask all the future positions, i.e., we set $\mathbf{Q}(j,j+1), \mathbf{Q}(j,j+1), \hdots, \mathbf{Q}(j,L)$ to $\mathrm{inf}$. This will avoid the softmax function from assigning dependency to events in the future.

In practice we stack multiple self-attention modules together, and inputs are passed through each of these modules sequentially. In this way our model is able to capture high level dependencies. We remark that stacking RNN/LSTM is not plausible because gradient explosion and gradient vanishing will render the stacked model difficult to train. Figure \ref{fig:THP-arch} illustrates the architecture of THP.

%%%%%%%%%%%%%%%%%%%%%%%%%%%%%%%%%%%%%%%%
\vspace{-0.05in}
\subsection{Continuous Time Conditional Intensity}
\vspace{-0.05in}

Dynamics of temporal point processes are described by a continuous conditional intensity function.  Eq.~\ref{eq:hidden-rep} only generates hidden representations for discrete time stamps, and the associated intensity is also discrete. Therefore an interpolated continuous time intensity function is in need.

Let $\lambda(t|\mathcal{H}_t)$ be the conditional intensity function for our model, where $\mathcal{H}_t=\{(t_j,k_j): t_j<t \}$ is the history up to time $t$. We define different intensity functions for different event types, i.e., for every $k \in \{1,2,\hdots,K\}$, define $\lambda_k(t|\mathcal{H}_t)$ as the conditional intensity function for events of type $k$. The conditional intensity function for the entire event sequence is defined by
\begin{equation} \nonumber
    \lambda(t|\mathcal{H}_t)=\sum_{k=1}^K \lambda_k(t|\mathcal{H}_t),  
\end{equation}
where each of the type-specific intensity takes the form
\begin{equation} \label{eq:intensity}
    \lambda_k(t|\mathcal{H}_t) = f_k \Big(
        \underbrace{\alpha_k \frac{t-t_j}{t_j}}_\textit{current} +
        \underbrace{\vphantom{\alpha_k \frac{t-t_j}{t_j}}\mathbf{w}_k^\top \mathbf{h}(t_j)}_\textit{history} +
        \underbrace{\vphantom{\alpha_k \frac{t-t_j}{t_j}}b_k}_\textit{base}
    \Big).
\end{equation}

In Eq.~\ref{eq:intensity}, time is defined on interval $t \in [t_j, t_{j+1})$, and $f_k(x) = \beta_k \log \big(1+\exp(x/\beta_k) \big)$ is the softplus function with ``softness'' parameter $\beta_k$. The reason for choosing this particular function is two-fold: first, the softplus function ensures that the intensity is positive; second, ``softness'' of the softplus function guarantees stable computation and avoids dramatic changes in the intensity.

Now we explain each term in Eq.~\ref{eq:intensity} in detail:

\noindent $\bullet$ The ``\textit{current}'' influence is an interpolation between two observed time stamps $t_j$ and $t_{j+1}$, and $\alpha_k$ modulates importance of the interpolation. When $t=t_j$, i.e., a new observation comes in, this influence is $0$. When $t \rightarrow t_{j+1}$, the conditional intensity function is no longer continuous. As a matter of fact, Eq.~\ref{eq:intensity} is continuous everywhere except for the observed events $\{(t_j,k_j)\}$. However, these ``jumps'' in intensity is a non-factor when computing likelihood. 

\noindent $\bullet$ The ``\textit{history}'' term contains two parts: a vector $\mathbf{w}_k$ that transforms the hidden states of the THP model into a scalar, and the hidden states $\mathbf{h}(t)$ (Sec.~\ref{sec:transformer-arch}) themselves that encode past events up to time $t$.
 
\noindent $\bullet$ The ``\textit{base}'' intensity represents probability of occurrence of events without considering history information.
% Notice that when $t$ is positive, the softplus function $f_k(t)$ is almost linear, and our formulation can decompose as $\lambda_k(t|\mathcal{H}_t) \approx f_k(\textit{base}) + f_k(\textit{current}+\textit{history})$.

With our proposed conditional intensity function, next time stamp prediction and next event type prediction is given by\footnote{Without causing any confusion, denote $\mathcal{H}_{t_j}$ as $\mathcal{H}_j$.}
\begin{equation} \label{eq:pred}
    \begin{split}
        & p(t|\mathcal{H}_t) = \lambda(t|\mathcal{H}_t) \exp \Big( -\int_{t_j}^t \lambda(\tau|\mathcal{H}_\tau) d\tau \Big), \\
        & \widehat{t}_{j+1} = \int_{t_j}^\infty t \cdot p(t|\mathcal{H}_t) dt,
        \quad \text{and }
        \widehat{k}_{j+1} = \underset{k}{\mathrm{argmax}} \ \frac{\lambda_k(t_{j+1}|\mathcal{H}_{j+1})}{\lambda(t_{j+1}|\mathcal{H}_{j+1})}.
    \end{split}
\end{equation}

%%%%%%%%%%%%%%%%%%%%%%%%%%%%%%%%%%%%%%%%
\vspace{-0.05in}
\subsection{Training} \label{sec:mle}
\vspace{-0.05in}

For any sequence $\mathcal{S}$ over an observation interval $[t_1,t_L]$, given its conditional intensity function $\lambda(t|\mathcal{H}_{t})$, the log-likelihood is
\begin{equation} \label{eq:ll}
    \ell(\mathcal{S}) =
        \underbrace{\sum_{j=1}^L \log \lambda(t_j|\mathcal{H}_{j})}_\text{event log-likelihood}
        - \underbrace{\vphantom{\sum_{j=1}^N \log \lambda(t_j|\mathcal{H}_{j})}\int_{t_1}^{t_L} \lambda(t|\mathcal{H}_{t}) dt}_\text{non-event log-likelihood}.
\end{equation}

Model parameters are learned by maximizing the log-likelihood across all sequences. Concretely, suppose we have $N$ sequences $\mathcal{S}_1, \mathcal{S}_2, \hdots, \mathcal{S}_N$ , then the goal is to find parameters that solve
\begin{equation} \nonumber
    \max~\sum_{i=1}^N \ell(\mathcal{S}_i),
\end{equation}
where $\ell(\mathcal{S}_i)$ is the log-likelihood of event sequence $\mathcal{S}_i$. This optimization problem can be efficiently solved by stochastic gradient type algorithms like ADAM \citep{adam}. Additionally, techniques that help stabilizing training such as layer normalization \citep{layer-norm} and residual connection \citep{residue} are also applied.

In Eq.~\ref{eq:ll}, one challenge is to compute $\Lambda=\int_{t_1}^{t_L} \lambda(t|\mathcal{H}_{t}) dt$, the non-event log-likelihood. Because of the softplus function, there is no closed-form computation for this integral, and a proper approximation is needed.

The first approach to approximate the non-event log-likelihood is by using Monte Carlo integration \citep{mcmc}:
\begin{equation} \label{eq:mcmc}
    \begin{split}
        &\widehat{\Lambda}_\text{MC} = \sum_{j=2}^L (t_j-t_{j-1}) \Big( \frac{1}{N} \sum_{i=1}^N \lambda(u_i) \Big), \quad
        \nabla \widehat{\Lambda}_\text{MC} = \sum_{j=2}^L (t_j-t_{j-1}) \Big( \frac{1}{N} \sum_{i=1}^N \nabla \lambda(u_i) \Big).
    \end{split}
\end{equation}
Here $u_i \sim \text{Unif}(t_{j-1},t_j)$ is sampled from a uniform distribution with support $[t_{j-1},t_j]$. Notice that $\lambda(u_i)$ and $\nabla \lambda(u_i)$ can be calculated by feed-forward and back-propagation through the model, respectively. Moreover, Eq.~\ref{eq:mcmc} yields an unbiased estimation to the integral, i.e., $\mathbb{E}[\widehat{\Lambda}_\text{MC}]=\Lambda$.

The second approach is to apply numerical integration methods, which are faster because of the elimination of sampling. For example, the trapezoidal rule \citep{numerical} states that
\begin{equation} \label{eq:trapezoidal}
    \widehat{\Lambda}_\text{NU} = \sum_{j=2}^L \frac{t_j-t_{j-1}}{2} \Big( \lambda(t_j | \mathcal{H}_{j})+\lambda(t_{j-1} | \mathcal{H}_{j-1}) \Big) 
\end{equation}
qualifies as an approximation to $\Lambda$. Other higher order methods such as the Simpson's rule \citep{numerical} can also be applied. Even though approximations build upon numerical integration algorithms are biased, in practice they are affordable. This is because the conditional intensity (Eq.~\ref{eq:intensity}) uses softplus as its activation function, which is highly smooth and ensures bias introduced by linear interpolations (Eq.~\ref{eq:trapezoidal}) between consecutive events are small.

%%%%%%%%%%%%%%%%%%%%%%%%%%%%%%%%%%%%%%%%%%%%%%%%%%%%%%%%%%%%%%%%%%%%%%%%%%%%%%%%
\vspace{-0.05in}
\section{Structured Transformer Hawkes Process} \label{sec:structure-thp}
\vspace{-0.05in}

\begin{figure}[tb!]
    \centering
    \includegraphics[width=0.6\linewidth]{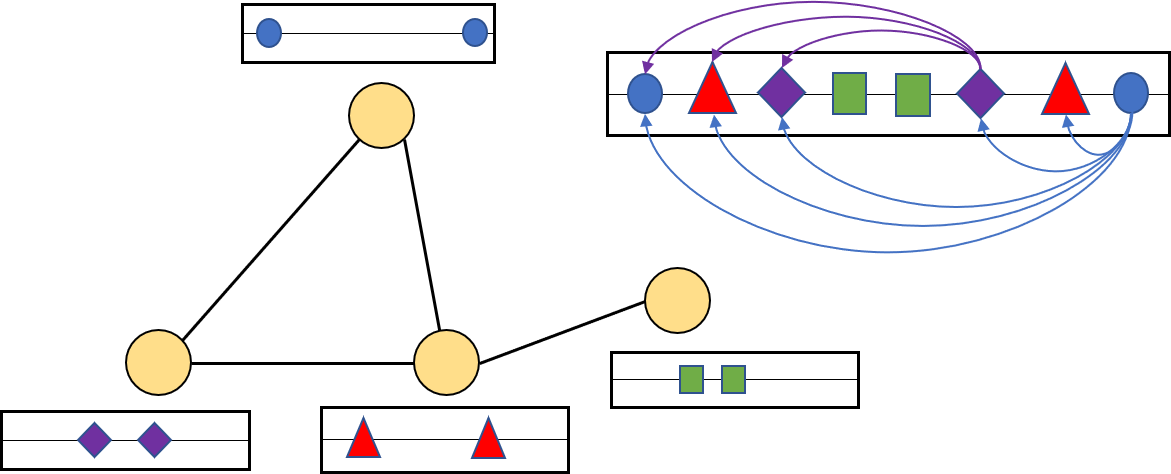}
    \vspace{-0.05in}
    \caption{Illustration of event sequences on a graph. Sequences on vertices are aligned temporally to form a long sequence, and relational information among events are shown in arrows. Notice that only the structural information of the last event (the blue circle) and the third to the last event (the purple diamond) are shown. Like before, events cannot attend to future.}
    \label{fig:graph-data}
\end{figure}

THP is quite general and can incorporate additional structural knowledge. We consider multiple point processes, where any two of them can be related. Such relationships are often described by a graph $\mathcal{G}=(\mathcal{V},\mathcal{E})$, where $\mathcal{V}$ is the vertex set, and each vertex is associated with a point process. Also, $\mathcal{E}$ is the edge set, where each edge signifies relational information between the corresponding two vertices. Figure \ref{fig:graph-data} illustrates event sequences on a graph.

The graph encodes relationships among vertices, and further indicates potential interactions. We propose to model all the point processes with a single THP, and the heterogeneity of the vertices’ point processes is handled by a vertex embedding approach.

Suppose we have an event sequence $\mathcal{S}=\{(t_j,k_j,v_j)\}_{j=1}^L$, where $t_j$ and $k_j$ are time stamps and event types as before. Further, $v_j \in \{1,2,\hdots,|\mathcal{V}|\}$ is an indicator to which vertex the event belongs. In addition to the event embedding and the temporal encoding (Eq.~\ref{eq:embedding}), we introduce a vertex embedding matrix $\mathbf{E} \in \mathbb{R}^{M \times |\mathcal{V}|}$, where the $j$-th column of $\mathbf{E}$ denotes the $M$-dimensional embedding for vertex $j$. Let $\mathbf{v}_j$ be the one-hot encoding of $v_j$, then embedding of $\mathcal{S}$ is specified by
\begin{equation} \nonumber
    \mathbf{X} = \big( \mathbf{U}\mathbf{Y} + \mathbf{E}\mathbf{V} + \mathbf{Z} \big)^\top,
\end{equation}
where $\mathbf{V}=[\mathbf{v}_1, \mathbf{v}_2, \hdots, \mathbf{v}_L] \in \mathbb{R}^{|\mathcal{V}| \times L}$ is the concatenation of vertices, and other terms are defined in Eq.~\ref{eq:embedding}.

The graph attention output is defined by
\begin{equation} \label{eq:STHP}
    \begin{split}
        & \mathbf{S} = \mathrm{Softmax} \left( \frac{\mathbf{Q}\mathbf{K}^\top}{\sqrt{M_K}} + \mathbf{A} \right) \mathbf{V_{\text{value}}}, \quad \text{where }
        \mathbf{A} = (\mathbf{E}\mathbf{V})^\top \mathbf{\Omega} (\mathbf{E}\mathbf{V}),
    \end{split}
\end{equation}
where $\mathbf{Q}$, $\mathbf{K}$, and $\mathbf{V}_{\text{value}}$ are the same\footnote{We use $\mathbf{V}_{\text{value}}$ to denote the value matrix instead of $\mathbf{V}$, which denotes the vertices.} as in Eq.~\ref{eq:attention}. Matrix $\mathbf{A} \in \mathbb{R}^{L \times L}$ is the vertex similarity matrix, where each entry $\mathbf{A}_{ij}$ signifies the similarity between two vertices $v_i$ and $v_j$, and $\mathbf{\Omega} \in \mathbb{R}^{M\times M}$ is a metric to be learned. To extend the graph self-attention module to a multi-head setting, we use different metric matrices $\{\mathbf{\Omega}_j\}_{j=1}^H$ for different heads.

We remark that unlike RNN-based shallow models, in structured-THP, multiple multi-head self-attention modules can be stacked (Figure \ref{fig:THP-arch}) to learn high level representations, a feature that enables learning of complicated similarities among vertices. Moreover, the vertex similarity matrix enables modeling of even more complicated structured data, such as sequences on dynamically evolving graphs.

With the incorporation of relational information, we need to modify the conditional intensity function accordingly. As an extension to Eq.~\ref{eq:intensity}, where each type of events has its own intensity, we define a different intensity function for each event type and each vertex. Specifically,
\begin{equation} \nonumber
    \begin{split}
        & \lambda(t|\mathcal{H}_t) = \sum_{k=1}^K \sum_{v=1}^{|\mathcal{V}|} \lambda_{k,v}(t|\mathcal{H}_t), \quad t \in [t_j, t_{j+1}), \quad \text{where }
        \lambda_{k,v}(t|\mathcal{H}_t) = f_{k,v} \Big( \alpha_{k,v} \frac{t-t_j}{t_j} + \mathbf{w}_{k,v}^\top \mathbf{h}(t) + b_{k,v} \Big).
    \end{split}
\end{equation}

Model parameters are learned by maximizing the log-likelihood (Eq.~\ref{eq:ll}) across all sequences. Concretely, suppose we have $N$ sequences $\mathcal{S}_1, \mathcal{S}_2, \hdots, \mathcal{S}_N$ , then parameters are obtained by solving
\begin{equation} \nonumber
    \max~\sum_{i=1}^N \ell(\mathcal{S}_i)  + \mu L_{\text{graph}}(\mathbf{V}, \mathbf{\Omega}),
\end{equation}
where $\mu$ is a hyper-parameter and
\begin{equation} \nonumber
    \begin{split}
        L_{\text{graph}}(\mathbf{V}, \mathbf{\Omega}) = & \sum_{k=1}^{|\mathcal{V}|} \sum_{j=1}^{k}
        -\log \big( 1+ \exp( \mathbf{V}_j \mathbf{\Omega} \mathbf{V}_k ) \big)
        + \mathbf{1}\{(v_j,v_k) \in \mathcal{E}\} \big( \mathbf{V}_j \mathbf{\Omega} \mathbf{V}_k \big).
    \end{split}
\end{equation}
Here $L_{\text{graph}}(\mathbf{V}, \mathbf{\Omega})$ is a regularization term that encourages $\mathbf{V}_j \mathbf{\Omega} \mathbf{V}_k$ to be large when there exists an edge between $v_j$ and $v_k$. Which means if two vertices are connected in graph $\mathcal{G}$, then the regularizer will promote attention between them, and vice versa.

Notice that in the simplest case, $\mathbf{A}$ in Eq.~\ref{eq:STHP} can be some transformation of the adjacency matrix, i.e., $\mathbf{A}_{ij}=1$ if $(v_i, v_j) \in \mathcal{E}$, and $0$ otherwise. However, we believe that this constraint is too strict, i.e., some connected vertices may not behave similarly. Therefore, we treat the graph as a guide and introduce a regularization term that encourages $\mathbf{A}$ to be similar to the adjacency matrix, but not enforce it. In this way, our model is more flexible.

%% file: experiment.tex
%!TEX root = main.tex

\section{Experiments} \label{sec:experiment}

We compare THP against existing models: Recurrent Marked Temporal Point Process (RMTPP,~\citet{rnn-hawkes}), Neural Hawkes Process (NHP,~\citet{neural-hawkes}), Time Series Event Sequence (TSES,~\citet{rnn-pp}), and Self-attentive Hawkes Processes (SAHP,~\citet{sahp})\footnote{This is a concurrent work that also employs the Transformer architecture, and we only include results reported in their paper.}. We evaluate the models by per-event log-likelihood (in nats) and event prediction accuracy on held-out test sets. Details about training are deferred to the appendix.

%%%%%%%%%%%%%%%%%%%%%%%%%%%%%%%%%%%%%%%%
\subsection{Datasets}

We adopt several datasets to evaluate the models. Table \ref{tab:dataset} summarizes statistics of the datasets.

\vspace{0.05in} \noindent $\bullet$
\textit{Retweets} \citep{hp-power}: The Retweets dataset contains sequences of tweets, where each sequence contains an origin tweet (i.e., some user initiates a tweet), and some follow-up tweets. We record the time and the user tag of each tweet. Further, users are grouped into three categories based on the number of their followers: ``small'', ``medium'', and ``large''.

\vspace{0.05in} \noindent $\bullet$
\textit{MemeTrack} \citep{snap}: This dataset contains mentions of $42$ thousand different memes spanning ten months. We collect data on over $1.5$ million documents (blogs, web articles, etc.) from over $5000$ websites. Each sequence in this dataset is the life-cycle of a particular meme, where each event (usage of meme) is associated with a time stamp and a website id.

\vspace{0.05in} \noindent $\bullet$
\textit{Financial Transactions} \citep{rnn-hawkes}: This financial dataset contains transaction records of a stock in one day. We record the time (in milliseconds) and the action that was taken in each transaction. The dataset is a single long sequence with only two types of events: ``buy'' and ``sell''. The event sequence is further partitioned by time stamps.

\vspace{0.05in} \noindent $\bullet$
\textit{Electrical Medical Records} \citep{mimic}: MIMIC-II medical dataset collects patients' visit to a hospital's ICU in a seven-year period. We treat the visits of each patient as a separate sequence, where each event in the sequence contains a time stamp and a diagnosis.

\vspace{0.05in} \noindent $\bullet$
\textit{StackOverflow} \citep{snap}: StackOverflow is a question-answering website. The website rewards users with badges to promote engagement in the community, and the same badge can be rewarded multiple times to the same user. We collect data in a two-year period, and we treat each user's reward history as a sequence. Each event in the sequence signifies receipt of a particular medal.

\vspace{0.05in} \noindent $\bullet$
\textit{911-Calls\footnote{The dataset is available on \url{www.kaggle.com/mchirico/montcoalert}.}}: The 911-Calls dataset contains emergency phone call records. Calling time, location of the caller, and nature of the emergency are logged for each record. We consider three types of emergencies: EMS, fire, and traffic. We treat location of callers (given by zipcodes) as vertices on a relational information graph. Zipcodes are ranked based on the number of recorded calls, and only the top $75$ zipcodes are kept. An undirected edge exists between two vertices if their zipcodes are within $10$ of each other.

\vspace{0.05in} \noindent $\bullet$
\textit{Earthquake\footnote{The dataset is provided by China Earthquake Data Center. (\url{http://data.earthquake.cn})}}: This dataset contains time and location of earthquakes in China in an eight-year period. We partition the records into two categories: ``small'' and ``large''. A relational information graph is built based on geographical locations of the earthquakes, i.e., each province is a vertex and earthquakes are sequences on the vertices. Two vertices are connected if their associated provinces are neighbors.

\begin{table}[t!]
\centering
\caption{Datasets statistics. From left to right columns: name of the dataset, number of event types, number of events in the dataset, and average length per sequence.}
\begin{tabular}{c|c|c|c}
    \hline
    Dataset & $K$ & \# events & Avg. length \\ \hline
    Retweets & $3$ & $2,173,533$ & $109$ \\
    MemeTrack & $5000$ & $123,639$ & $3$ \\
    Financial & $2$ & $414,800$ & $2074$ \\
    MIMIC-II & $75$ & $2,419$ & $4$ \\
    StackOverflow & $22$ & $480,413$ & $72$ \\
    911-Calls & $3$ & $290,293$ & $403$ \\ 
    Earthquake & $2$ & $256,932$ & $500$ \\ \hline
\end{tabular}
\label{tab:dataset}
\end{table}

%%%%%%%%%%%%%%%%%%%%%%%%%%%%%%%%%%%%%%%%
\subsection{Training Details}

To facilitate comparison with previous works, all the datasets are used by \citet{rnn-hawkes} and \citet{neural-hawkes}, except for 911-Calls and Earthquake. Details about data pre-processing and train-dev-test split, as well as downloadable links, can be found in the aforementioned papers. For the 911-Calls dataset, we exclude zipcodes (and the associated events) whose occurrences are scarce, i.e., we only keep zipcodes that have the top $75$ frequent occurrences. The dataset contains $141$ types of events, and we cluster them into three categories, namely EMS, fire, and traffic. We do not exclude any events in the Earthquake dataset. Earthquakes are partitioned into two categories, ``small'' and ``large'', where small earthquakes are the ones whose Richter scale is equal to or lower than $1.0$. We perform this partition because of the imbalance in data, i.e., most of the recorded earthquakes are on small magnitude. Models are trained on 911-Calls and Earthquake with different number of training events. In each experiment, we equally divide the events that are not in the training set in half to construct the development set and the test set.

There are three sets of hyper-parameters that we use, and they are summarized in Table~\ref{tab:hyper-params}. Besides layer normalization and residual connection, we also employ the dropout technique to avoid overfitting. Table~\ref{tab:dataset-params} contains the specific parameters that are applied for the training of each dataset. In the table, from left to right columns specify: name of the dataset, the set of applied hyper-parameters, batch size, learning rate, and solver for the integral approximation (MC stands for Monte Carlo integration, and NU stands for numerical integration with the trapezoidal rule), respectively. In the 911-Calls and the Earthquakes datasets, we also employ the graph regularization method, and the corresponding regularization parameter is set to be $0.01$ in all the experiments. We use a single NVIDIA RTX graphics card to run all the experiments.

\begin{table}[t!]
    \centering
    \caption{Sets of hyper-parameters used in training.}
    \begin{tabular}{c|cccccc}
        \hline
         Parameters & \# head & \# layer & $M$ & $M_K=M_V$ & $M_H$ & dropout \\ \hline
         Set 1 & $3$ & $3$ & $64$ & $16$ & $256$ & $0.1$ \\
         Set 2 & $6$ & $6$ & $128$ & $64$ & $2048$ & $0.1$ \\
         Set 3 & $4$ & $4$ & $512$ & $512$ & $1024$ & $0.1$ \\ \hline
    \end{tabular}
    \label{tab:hyper-params}
\end{table}

\begin{table}[t!]
    \centering
    \caption{Hyper-parameters used for training each dataset.}
    \begin{tabular}{c|ccccccc}
        \hline
        Dataset & Retweets & Meme & Financial & MIMIC & StackOverflow & 911 & Earthquake \\ \hline
        Set & 1 & 1 & 2 & 1 & 3 & 2 & 3 \\
        Batch & $16$ & $128$ & $1$ & $1$ & $4$ & $1$ & $1$ \\
        LR & $1 \times 10^{-2}$ & $1 \times 10^{-3}$ & $1 \times 10^{-4}$ & $1 \times 10^{-4}$ & $1 \times 10^{-4}$ & $1 \times 10^{-5}$ & $1 \times 10^{-5}$ \\
        Solver & MC & MC & NU & NU & NU & MC & MC \\ \hline
    \end{tabular}
    \label{tab:dataset-params}
\end{table}

%%%%%%%%%%%%%%%%%%%%%%%%%%%%%%%%%%%%%%%%
\subsection{Likelihood Comparison}

\begin{figure}[t!]
    \centering
    \includegraphics[width=0.8\linewidth]{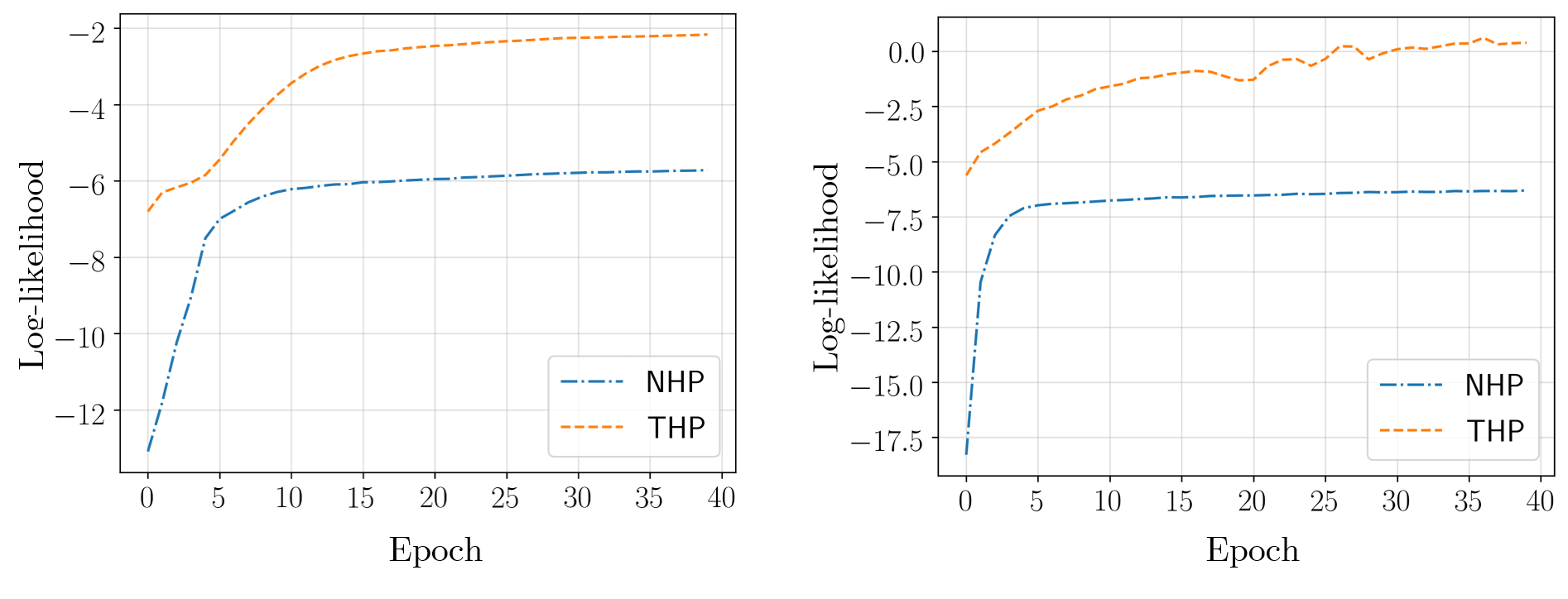}
    \vskip -0.2in
    \caption{Training curves of NHP and THP fitted on Retweets (left figure) and MemeTrack (right figure). Validation log-likelihood is (NHP vs. THP): -$5.60$ vs. -$2.04$ for Reweets, and -$6.23$ vs. $0.68$ for MemeTrack.}
    \label{fig:ll}
\end{figure}

We fit THP and NHP on Retweets and MemeTrack. From Figure \ref{fig:ll}, we can see that THP outperforms NHP during the entire training process by large margins on both of the datasets. The reason is because of the complicated nature of social media data, and RNN-based models such as NHP are not powerful enough to model the dynamics.

In the Retweets dataset, we often observe time gaps between two consecutive retweets become larger, and this dynamic can be successfully modeled by temporal encoding. Also, unlike RNN-based models, our model is able to capture long-term dependencies that exist in long sequences. In the MemeTrack dataset, we have extremely short sequences, i.e., average sequence length is $3$. Even though the data only exhibit short-term dependencies, we still need to model latent properties of memes such as topics and targeted users. We build deep THP models to capture these high-level features, and we remark that constructing deep NHP is not plausible because of the difficulty in training.

Table \ref{tab:ll} summarizes results on other datasets. Note that TSES is likelihood-free. Our THP model fits the data well and outperforms all the baselines in all the experiments.

\begin{figure}[t!]
    \centering
    \includegraphics[width=1.0\linewidth]{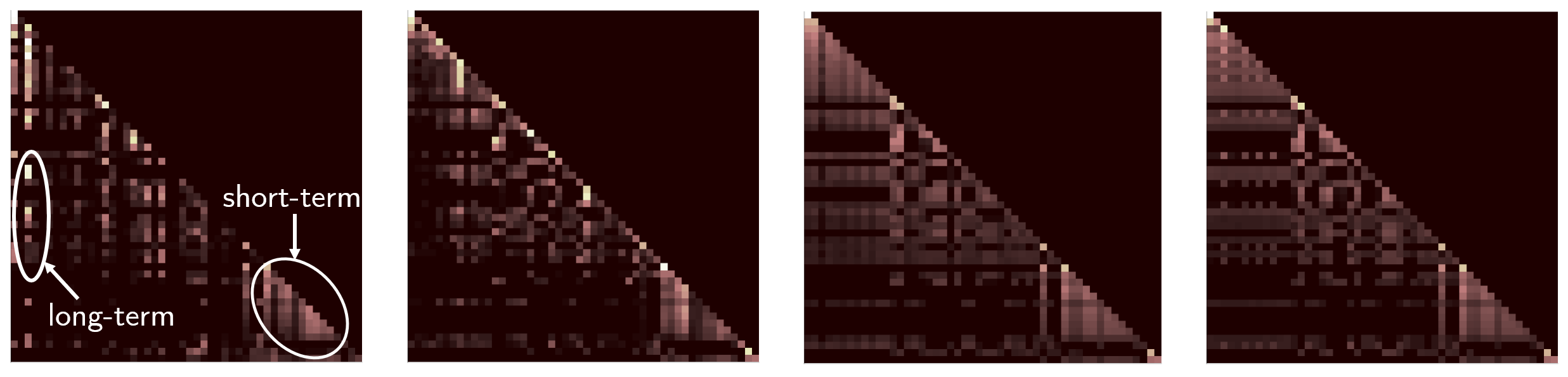}
    \vskip -0.2in
    \caption{Visualization of attention patterns of different attention heads in different layers. Pixel $(i,j)$ in each figure signifies the attention weight of event $(t_j,k_j)$ attending to event $(t_i,k_i)$. Attention heads in the upper two figures are from the first layer, while they are from the last layer in the lower two figures.}
    \label{fig:attention}
\end{figure}

Figure \ref{fig:attention} visualizes attention patterns of THP. We can see that each attention head employs a different pattern to capture dependencies. Moreover, while attention heads in the first layer tend to focus on individual events, the attention patterns in the last layer are more uniformly distributed. This is because features in deeper layers are already transformed by attention heads in shallow layers.

\begin{table}[t!]
\centering
\caption{Log-likelihood comparison.}
\begin{tabular}{c|ccccc}
\hline
Model & Retweets    & MemeTrack    & Financial   & MIMIC-II & StackOverflow    \\ \hline
RMTPP & -5.99 & -6.04 & -3.89 & -1.35 & -2.60 \\
NHP   & -5.60 & -6.23 & -3.60 & -1.38 & -2.55 \\
SAHP  & -4.56 & ---   & ---   & -0.52 & -1.86 \\
THP   & \textbf{-2.04} & \textbf{0.68}  & \textbf{-1.11} & \textbf{0.820} & \textbf{0.042} \\ \hline
\end{tabular}
\label{tab:ll}
\end{table}

%%%%%%%%%%%%%%%%%%%%%%%%%%%%%%%%%%%%%%%%
\subsection{Event Prediction Comparison}

For point processes, event prediction is just as important as data fitting. Eq.~\ref{eq:pred} enables us to predict future events. In practice, however, adding additional prediction layers on top of the THP model yields better performance. Specifically, given the hidden representation $\mathbf{h}(t_j)$ for event $(t_j,k_j)$, the next event type and time predictions are as follows.

\noindent $\bullet$ The next event type prediction is
\begin{equation} \nonumber
    \begin{split}
        \widehat{k}_{j+1} = \underset{k}{\mathrm{argmax}}~\widehat{\mathbf{p}}_{j+1}(k),
        \quad \text{where }
        \widehat{\mathbf{p}}_{j+1} = \mathrm{Softmax} \big(\mathbf{W}^\text{type} \mathbf{h}(t_j) \big),
    \end{split}
\end{equation}
where $\mathbf{W}^\text{type} \in \mathbb{R}^{K \times M}$ is the predictor parameter, and $\widehat{\mathbf{p}}_j(k)$ is the $k$-th element of $\widehat{\mathbf{p}}_j \in \mathbb{R}^{K}$.

\noindent $\bullet$ The next event time prediction is
\begin{equation} \nonumber
    \widehat{t}_{j+1} = \mathbf{W}^\text{time} \mathbf{h}(t_j),
\end{equation}
where $\mathbf{W}^\text{time} \in \mathbb{R}^{1 \times M}$ is the predictor parameter.

\begin{table}[t!]
\begin{minipage}[t]{0.49\textwidth}
    \centering
    \caption{Event type prediction accuracy comparison. Here FIN is the Financial Transactions dataset, and SO is the StackOverflow dataset.}
    \begin{tabular}{c|ccc}
        \hline
        Model & FIN & MIMIC-II & SO \\ \hline
        RMTPP & 61.95     & 81.2     & 45.9          \\
        NHP   & 62.20     & 83.2     & 46.3          \\
        TSES  & 62.17     & 83.0     & 46.2          \\
        THP   & \textbf{62.64}     & \textbf{85.3}     & \textbf{47.0}          \\ \hline
    \end{tabular}
    \label{tab:type}
\end{minipage} \qquad
\begin{minipage}[t]{0.49\textwidth}
    \centering
    \caption{Event time prediction RMSE comparison. Here FIN is the Financial Transactions dataset, and SO is the StackOverflow dataset.}
    \begin{tabular}{c|ccc}
        \hline
        Model & FIN & MIMIC-II & SO \\ \hline
        RMTPP & 1.56      & 6.12     & 9.78          \\
        NHP   & 1.56      & 6.13     & 9.83          \\
        TSES  & 1.50      & 4.70     & 8.00          \\
        SAHP  & ---       & 3.89     & 5.57          \\
        THP   & \textbf{0.93}      & \textbf{0.82}     & \textbf{4.99}          \\ \hline
    \end{tabular}
    \label{tab:time}
\end{minipage}
\end{table}

\begin{figure}[t!]
    \centering
    \includegraphics[width=0.7\linewidth]{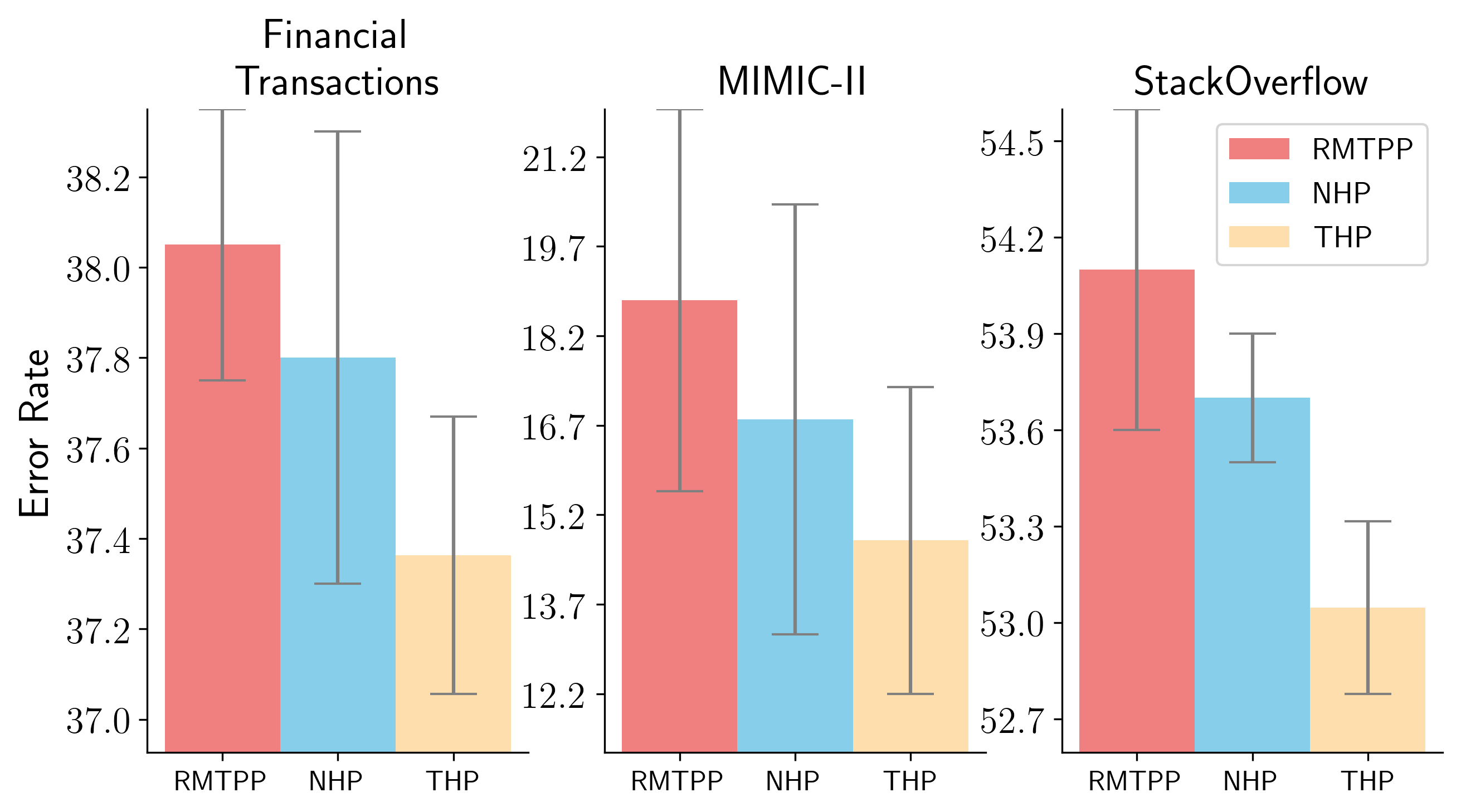}
    \vskip -0.2in
    \caption{Prediction error rates of THP, NHP, and RMTPP. Based on a same train-dev-test splitting ratio, each dataset is sampled five times to produce different train, development and test sets. Error bars are generated according to these experiments.}
    \label{fig:pred-acc}
\end{figure}

To learn the predictor parameters, the loss function is equipped with a cross-entropy term for event type prediction and a mean square error term for event time prediction. Concretely, for an event sequence $\mathcal{S}=\{(t_j,k_j)\}_{j=1}^L$, let $\mathbf{k}_1, \mathbf{k}_2, \hdots, \mathbf{k}_L$ be the ground-truth one-hot encodings for the event types, we define
\begin{equation} \nonumber
    \begin{split}
        & L_\text{type}(\mathcal{S}) = \sum_{j=2}^L - \mathbf{k}_j^\top \log(\widehat{\mathbf{p}}_j),
        \quad \text{and }
        L_\text{time}(\mathcal{S}) = \sum_{j=2}^L (t_j - \widehat{t}_j)^2.
    \end{split}
\end{equation}
Notice that we do not predict the first event. Then, given event sequences $\{\mathcal{S}_i\}_{i=1}^N$, we seek to solve
\begin{equation} \nonumber
    \min~\sum_{i=1}^N -\ell(\mathcal{S}_i) + L_\text{type}(\mathcal{S}_i) + L_\text{time}(\mathcal{S}_i),
\end{equation}
where $\ell(\mathcal{S}_i)$ is the log-likelihood (Eq.~\ref{eq:ll}) of $\mathcal{S}_i$.

To evaluate model performance, we predict every held-out event $(t_j,k_j)$ given its history $\mathcal{H}_j$, i.e., for a test sequence of length $L$, we make $L-1$ predictions.
We evaluate event type prediction by accuracy and event time prediction by Root Mean Square Error (RMSE).
Table \ref{tab:type} and Table \ref{tab:time} summarize experiment results. We can see that THP outperforms the baselines in all these tasks. The datasets we adopted vary significantly in average sequence length, i.e., the average length in Financial Transactions is $2074$ while it is only $4$ in MIMIC-II. In all the three datasets, THP improves upon RNN-based models by a notable margin. The results demonstrate that THP is able to capture both short-term and long-term dependencies better than existing methods.

Figure \ref{fig:pred-acc} illustrates run-to-run variance of THP, NHP, and RMTPP. The error bars are wide because of how the data are split. Held-out test sets are constructed by randomly sampling some events from the entire dataset. That is, at times “important” events are sampled out, which will yield unsatisfactory model performance. Our results are better than all the baselines in all the individual experiments.

%%%%%%%%%%%%%%%%%%%%%%%%%%%%%%%%%%%%%%%%
\subsection{THP vs. Structured-THP}

\begin{figure}[t!]
    \centering
    \includegraphics[width=0.75\linewidth]{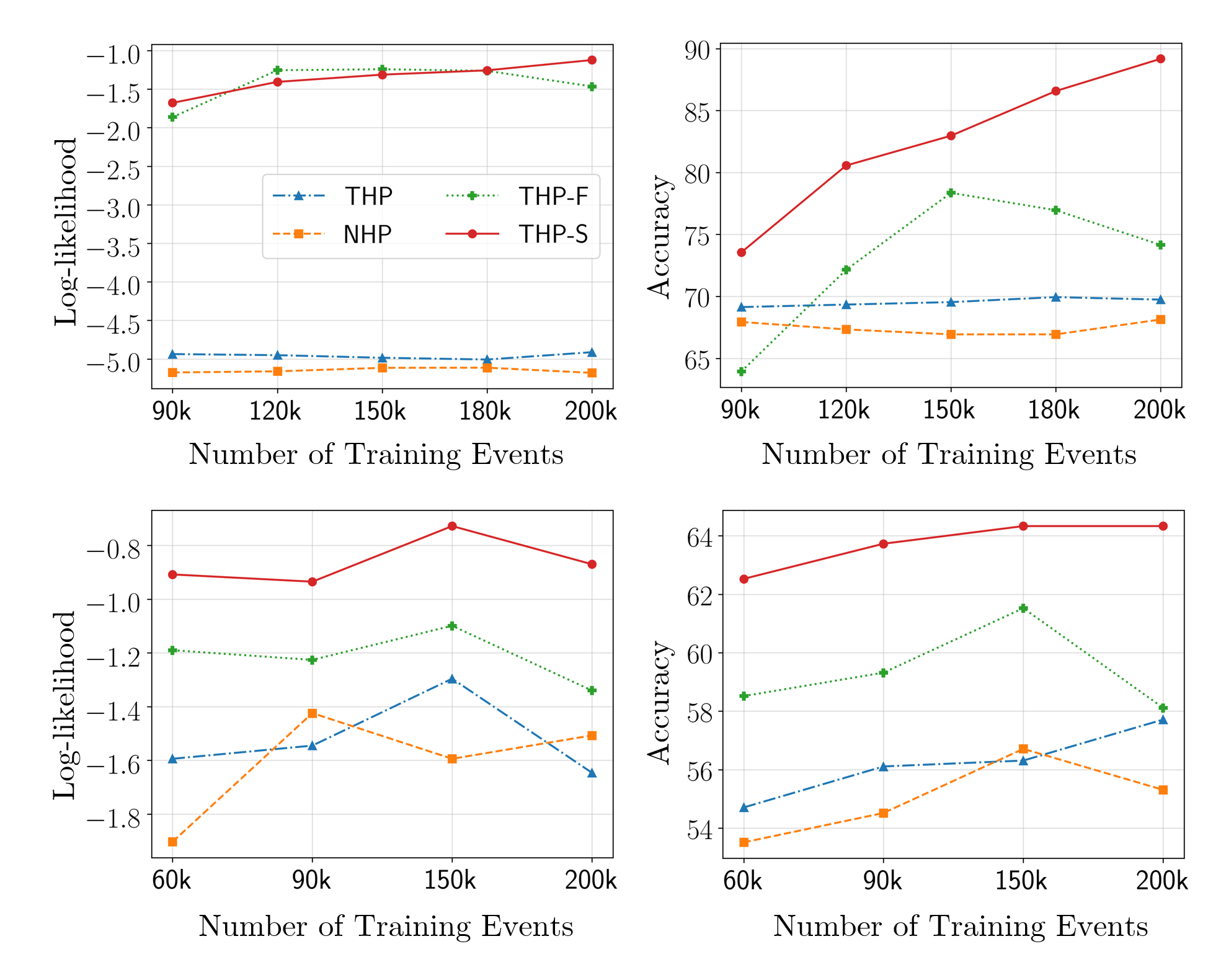}
    \vskip -0.2in
    \caption{Log-likelihood and prediction accuracy of NHP, THP, THP with full attention (THP-F), and structured-THP (THP-S) fitted on the 911-Calls (upper two figures) and the Earthquake (lower two figures) datasets. Models are trained using different number of events.}
    \label{fig:911}
\end{figure}

Now we demonstrate by incorporating relational information, THP achieves improved performance. 

Baseline models are constructed as following: for each vertex on a relational graph $\mathcal{G}$, there exists a point process that consists of time and type of events. These event sequences are learned separately by both THP and NHP, i.e., we do not allow information sharing among vertices in these models.

To integrate $\mathcal{G}$ into THP, we consider two approaches. The first approach is by allowing full attention, i.e., information from one vertex can be shared with all the other vertices. The second approach is by using the neighborhood graph, which is constructed based on spatial proximity. In this approach, a specific vertex can only share information with its neighbors. We fit a structured-THP to both of the cases.

Figure \ref{fig:911} summarizes experimental results.
We can see that THP is comparable or better than NHP in both validation likelihood and event prediction, which further demonstrates that THP can model complicated dynamics better than RNN-based models.
Notice that THP-F, the structured-THP with full attention, yields a much better likelihood than the baseline models, which means relational information sharing can help the models in capturing latent dynamics.
However, unlike likelihood, THP-F does not show consistent improvements in event prediction. This is because when the number of training events is small, the model cannot build a sufficient information-sharing heuristic. Also, the performance drop when the number of training events is large is due to the inhomogeneity of data. This demonstrates that the full attention scheme results in undesirable dependencies on which the attention heads focus.
THP-S successfully resolves this issue by eliminating such dependencies from the attention heads' span based on spatial closeness of vertices. In this way, THP-S further improves upon THP-F, especially in event prediction tasks.

%%%%%%%%%%%%%%%%%%%%%%%%%%%%%%%%%%%%%%%%
\subsection{Ablation Study}

\begin{table}[b!]
\vskip -0.15in
\begin{minipage}[t]{0.45\textwidth}
    \caption{Log-likelihood of variants of NHP and THP fitted on Retweets and MemeTrack. TE stands for temporal encoding (Eq.~\ref{eq:temp-encoding}), and PE stands for positional encoding \citep{transformer}.}
    \centering
    \begin{tabular}{l|cc}
        \hline
         Model & Retweets & MemeTrack  \\ \hline
         NHP & $-5.60$ & $-6.23$ \\
         NHP + TE & $-2.50$ & $-1.64$ \\ \hline
         Atten & $-5.29$ & $-5.09$ \\
         Atten + PE & $-5.25$ & $-4.70$ \\
         Atten + TE & $\mathbf{-2.03}$ & $\textbf{0.68}$ \\ \hline
    \end{tabular}
    \label{tab:ablation}
\end{minipage} \hspace{0.2in}
\begin{minipage}[t]{0.45\textwidth}
    \caption{Sensitivity to the number of parameters and run-time comparison. Speedup is the speed of THP against NHP.}
    \centering
    \begin{tabular}{c|cc|c}
        \hline
        \multirow{2}{*}{\# parameters} & \multicolumn{2}{c|}{Log-likelihood} & \multirow{2}{*}{Speedup} \\ \cline{2-3}
                                       & THP              & NHP              &                          \\ \hline
        $100$k & $-2.090$ & $-6.019$  & $\times 1.985$ \\
        $200$k & $-2.072$ & $-5.595$  & $\times 2.564$ \\
        $500$k & $-2.058$ & $-5.590$ & $\times 2.224$ \\
        $1000$k & $-2.060$ & $-5.614$ & $\times 1.778$ \\ \hline
    \end{tabular}
    \label{tab:sense}
\end{minipage}
\end{table}

We perform ablation study on Retweets and MemeTrack, and we evaluate models by validation log-likelihood. We inspect variants of THP by removing the self-attention and the temporal encoding mechanisms. Moreover, we test the effect of temporal encoding on NHP. Table \ref{tab:ablation} summarizes experimental results. As shown, both the self-attention module and the temporal encoding contribute to model performance.

We examine the models' sensitivity to the number of parameters on the Retweets dataset. As shown in Table \ref{tab:sense}, our model is not sensitive to its number of parameters. Without the recurrent structure, Transformer-based models often have large number of parameters, but our THP model can outperform RNN-based models with fewer parameters. In all the experiments, using a small model (about 100-200k parameters) will suffice. In comparison, NHP has about 1000k and TSES has about 2000k parameters to achieve the best performance, which are much larger than THP. We also include run-time comparison in Table \ref{tab:sense}. We conclude that THP is efficient in both model size and training speed.

%% file: conclusion.tex
%!TEX root = main.tex

\section{Conclusion} \label{sec:conclusion}

In this paper we present Transformer Hawkes Process, a framework for analyzing event streams. Event sequence data are common in our daily life, and they exhibit sophisticated short-term and long-term dependencies. Our proposed model utilizes the self-attention mechanism to capture both of these dependencies, and meanwhile enjoys computational efficiency. Moreover, THP is quite general and can integrate structural knowledge into the model. This facilitates analyzing more complicated data, such as event sequences on graphs. Experiments on various real-world datasets demonstrate that THP achieves state-of-the-art performance in terms of both likelihood and event prediction accuracy.